\documentclass[letterpaper, 10 pt, conference]{ieeeconf}  

\usepackage{graphicx}
\usepackage{amsmath,amssymb,amsfonts}

\usepackage{textcomp}
\usepackage[hyphens]{url}
\usepackage{xurl}
\usepackage{microtype}

\usepackage{booktabs}
\usepackage{algorithm}    
\usepackage{algpseudocode} 

\usepackage{balance}

\IEEEoverridecommandlockouts                              

\title{\LARGE \bf
Generative Modeling for Adversarial Lane-Change Scenarios
}
\author{Chuancheng Zhang$^{1,2 \dag}$, Zhenhao Wang$^{3 \dag}$,  Jiangcheng Wang$^{4}$, Kun Su$^{2}$, Qiang Lv$^{2}$, Bin Jiang$^{1,2}$, \\Kunkun Hao$^{4*}$, Wenyu Wang$^{2*}$
\thanks{\dag Both authors contributed equally to this research.}
\thanks{*Corresponding author email: haokunkun@synkrotron.ai}
}

\begin{document}

\maketitle
\thispagestyle{empty}
\pagestyle{empty}

\begin{abstract}
Decision-making in long-tail scenarios is pivotal to autonomous-driving development, and realistic and challenging simulations play a crucial role in testing safety-critical situations. However, existing open-source datasets lack systematic coverage of long-tail scenes, and lane-change maneuvers being emblematic, rendering such data exceedingly scarce. To bridge this gap, we introduce a data mining framework that exhaustively analyzes two widely used datasets, NGSIM and INTERACTION, to identify sequences marked by hazardous behavior, thereby replenishing these neglected scenarios. Using Generative Adversarial Imitation Learning (GAIL) enhanced with Proximal Policy Optimization (PPO), and enriched by vehicular-environment interaction analytics, our method iteratively refines and parameterizes newly generated trajectories. Distinguished by a rationally adversarial and sensitivity-aware perspective, the approach optimizes the creation of challenging scenes. Experiments show that, compared to unfiltered data and baseline models, our method produces behaviors that are simultaneously both adversarial and natural, judged by collision frequency, acceleration profiles, and lane-change dynamics, offering constructive insights to amplifying long-tailed lane-change instances in datasets and advancing decision-making training. The video demo of the evaluation process can be found at: \url{https://www.youtube.com/watch?v=RoyfG_B-EGw}

\end{abstract}


\begin{figure*}[ht]
\centering
\includegraphics[width=1.0\textwidth, trim=0 60 0 0, clip]{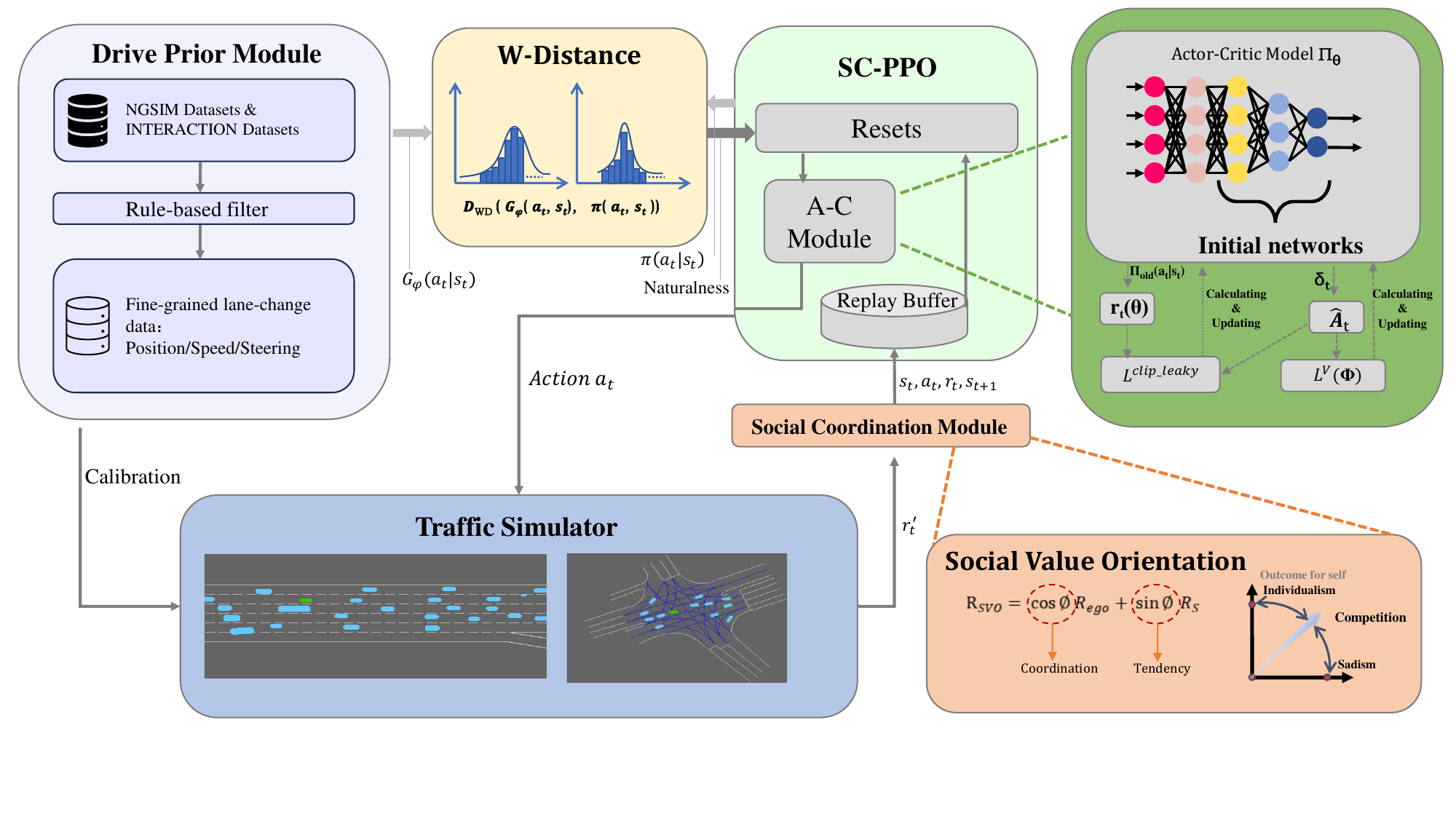} 
\caption{The overall framework of our sensitivity and continuity scenario generation solution.}
\label{icml-historical}
\vspace{-0.2in}
\end{figure*}

\section{INTRODUCTION}
In long-tail scenarios such as complex environments, emergencies, and extreme conditions, the lack of sufficient historical data limits the ability of autonomous driving system to respond to these situations, making it difficult to effectively predict and mitigate associated risks \cite{c1}. Lane-change maneuvers, one of the most fundamental yet complex driving behaviors, is a key aspect of long-tail scenarios that challenge autonomous systems \cite{c2}. These maneuvers involve dynamic interactions among multiple vehicles across adjacent lanes. This behavior exhibits significant variability, particularly in highway and urban traffic contexts: random lane changes are more common on highways and can disrupt traffic flow, thereby reducing safety \cite{c3}; while forced lane changes are primarily observed in busy urban sections, potentially leading to reduced lane capacity and generating shockwave effects \cite{c4}. In real life, lane-change maneuvers are often associated with different types of collisions, such as rear-end and side-swipe accidents. For example, in 2019, New South Wales, Australia, reported 830 lane-change collision incidents (TfNSW, 2020), and in the same year, lane-change collisions accounted for 3\% of the total collision incidents in Queensland, Australia (DTMR, 2020). In the United States, side-swipe accidents constituted 13\% of total collisions in 2019 (NHTSA, 2020), \cite{c5}. These statistics underscore that the risks associated with lane-changing behaviors cannot be overlooked and that a deep understanding of lane-change decision-making and interaction processes is essential.

The growth of data volume can significantly enhance model performance; however, once the data reaches a certain volume, the growth in performance tends to plateau. Moreover, autonomous vehicles (AVs) will inevitably encounter scenarios that are not present in the training data \cite{c6}. From this perspective, it becomes clear that, in the field of autonomous driving, simply expanding the data volume is not always necessary. For the vast majority of traffic scenarios, it is not essential to have an extremely large dataset to achieve coverage; instead, the focus should shift from simply expanding data to collecting targeted safety-critical scenario data, which is more critical to ensuring the robustness of the autonomous driving system. Thus, this motivates our approach: to represent roadside-collected data with existing open-source datasets, refine and filter safety-critical scenario data, and then generate these scenarios through artificial intelligence methods, thereby supplementing the data for very rare traffic scenarios. 
Based on two datasets, NGSIM~\cite{c7} and INTERACTION~\cite{c8}, we develop a rule-based approach to mine potential hazardous scenarios (Drive Prior Module in Fig.~\ref{icml-historical}. This approach employs a deep reinforcement learning (DRL) framework that incorporates both adversarial and natural characteristics to generate highway and urban traffic flow data with candidate lane-change safety-critical scenarios.
GAIL, a variant of inverse reinforcement learning, has proven effective in generating safety-critical scenarios~\cite{c9}. Unlike traditional inverse reinforcement learning methods, GAIL learns directly from refined expert trajectories without requiring explicit reward function modeling. Compared to diffusion models, GAIL generates more precise scenarios in small-sample distributions by modeling specific behaviors instead of the entire data distribution. By introducing the Leaky and Resets techniques \cite{c11, c12}, we significantly increased the model's sensitivity and capacity for sustainable learning. Furthermore, as shown in Fig.~\ref{icml-historical}, we incorporate the Social Value Orientation (SVO) mechanism \cite{c13} to enhance the exploration capability of adversarial vehicles in scenario generation, ensuring that the generated scenario data strikes a balance between safety and naturalness \cite{c14}. Through refined data mining and model optimization for the corresponding data, our approach generates realistic safety-critical scenarios more efficiently, leading to more authentic adversarial behaviors compared to baseline models. This innovative method not only provides a feasible solution for supplementing hazardous scenarios but also lays the groundwork for enhancing the safety of future autonomous driving systems. The contributions of this work are as follows:
\begin{enumerate}
\item A rule-based pipeline labels hazardous driving and yields authentic safety-critical scenarios from large-scale datasets for mass production, alleviating per-distribution sample scarcity in expert-collected data.
\item To balance naturalness and adversarial robustness, we propose an SVO-based reward function that models surrounding-vehicle influences on the ego, encouraging alternative planned trajectories and generating diverse new scenarios.
\item The SCPPO (Sensitivity and Continuity) algorithm is integrated into GAIL as the generator to enhance its long-term learning capability and sensitivity to driving behaviors, enabling finer-grained action exploration.
\end{enumerate}

\section{Related Works}
\subsection{Refined Mining of Data in Existing Open Source Datasets}
In recent years, with the advancement of autonomous driving research, refined data mining in open-source datasets has become a crucial direction for enhancing model performance. Through in-depth data processing and optimization, key features in driving interaction behaviors can be more effectively captured, which in turn improves the model’s ability to generalize across varied scenarios. For example, Cheng et al. \cite{c15} reduced composite errors in the nuPlan dataset using data augmentation techniques, subsequently developing a powerful baseline model. In mining the NGSIM dataset, Zhou et al. \cite{c16} integrated and deeply explored the NGSIM dataset using the SMARTS platform, extracting 3366 vehicle trajectories and employing PPO to train a reinforcement learning model, demonstrating superior performance in reducing hazardous events. Furthermore, Li et al. \cite{c17} first filtered the NGSIM data and then paired it with a Transformer to improve the accuracy of trajectory prediction. However, a key limitation is the simplistic nature of the data filtering process, which often overlooks the latent complexities within the data, leaving critical interactions unexplored. In contrast, Jiang et al. \cite{c18} conducted a more profound analysis of the INTERACTION dataset, extracting a dataset with high-density interaction behaviors.

\subsection{Data-driven Scenario Generation}
Diffusion-based generative models have been explored for scenario synthesis, offering advantages in capturing complex, high-dimensional behavior distributions.
Wang et al.~\cite{c34} reframed 3D occupancy prediction as a generative modeling problem, showing that diffusion models capture complex 3D structures, handle noisy and incomplete labels, and better represent multimodal occupancy distributions compared to discriminative baselines. 
Scenario diffusion, proposed by Pronovost et al.~\cite{c35}, is a diffusion-based architecture for the generation of controllable traffic scenarios. It integrates latent diffusion with map- and token-conditioned generation to produce agent bounding boxes and trajectories, enabling control over global and local scene properties. This facilitates targeting rare or safety-critical scenarios, with experiments confirming generalization across regions. Rempe et al.~\cite{c36} further proposed the generation of useful accidents-prone scenarios through a learned traffic prior, enabling generative models to capture higher-level semantic regularities in risky driving behaviors.


DRL-based methods are already capable of effectively generating simple, discrete adversarial traffic scenarios and controlling car-following behavior in the presence of surrounding vehicles \cite{c19, c20}. Furthermore, some studies employ Deep Deterministic Policy Gradient (DDPG) to control surrounding agents for lane-change scenario generation and the implementation of adversarial strategies \cite{c21, c22, c23}. Wachi et al. \cite{c24} also adopted a DDPG-based approach for multi-agent control of surrounding vehicles, creating scenarios that more closely resemble real-world conditions. 
Chen et al.~\cite{c37} introduced FREA, a feasibility-guided adversarial scenario generation framework that emphasizes both safety-criticality and behavioral plausibility, ensuring generated scenarios remain adversarial yet realistic. He et al.~\cite{c33} proposed R-DDPG, a constrained DDPG framework with a rationality reward to penalize unrealistic accelerations, training adversarial agents to generate realistic yet collision-prone scenarios. This method effectively balances scenario realism and adversarial challenge in ego-vehicle safety evaluation. 


\section{Experiment Design and Methodology}
\subsection{Datasets and Data Preprocessing}
\begin{figure}[ht]
\begin{center}
\centerline{\includegraphics[width=0.95\columnwidth, trim=10 10 10 10, clip]{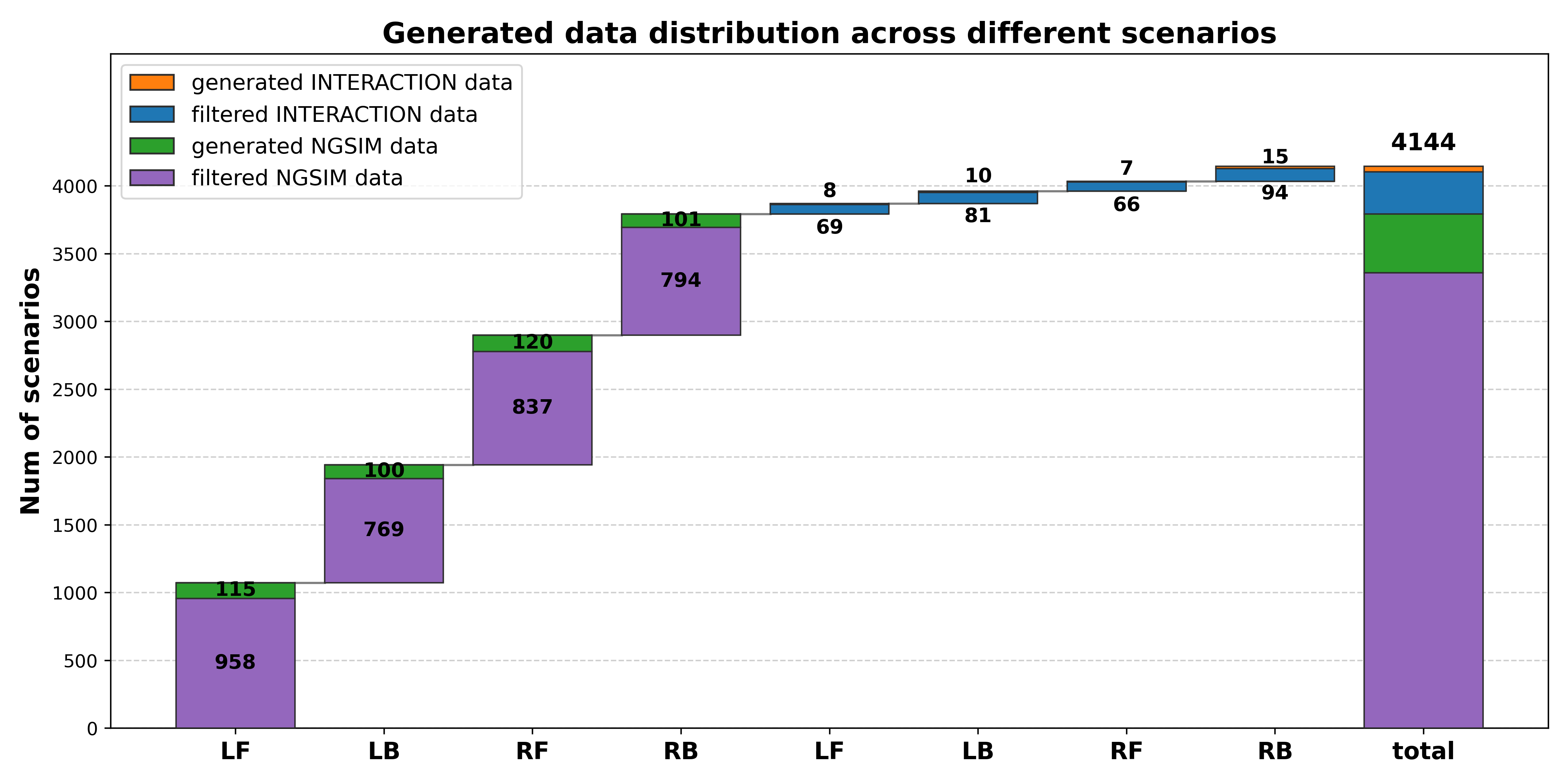}}
\caption{The distribution and comparison of data regenerated based on filtering in different scenarios. Among them, \textbf{LF} represents left front with adversarial vehicles, \textbf{LB} represents left behind with adversarial vehicles, \textbf{RF} represents right front with adversarial vehicles, and \textbf{RB} represents right behind with adversarial vehicles. The scenario conversion success rate is approximately 13\% (476:3668).}
\label{fig:data_compare}
\end{center}
\end{figure}

A refined data mining process is applied to the NGSIM and INTERACTION datasets to identify dangerous behaviors and address the lack of safety-critical scenarios. Lane change events are extracted based on map data and vehicle pose sequences. For NGSIM, we constructed a 5-lane highway; for INTERACTION, we used OSM maps to match vehicle poses to lane indices. A lane change is detected when the current lane index differs from the previous one, with both indices within the nearby lane list of the main vehicle. Corresponding frames are recorded, and surrounding vehicles in both original and new lanes are tracked. 
Algorithm~\ref{alg:lane_change_extraction} provides an overview of the lane change scenario extraction. Subsequently, a systematic cleaning and normalization process is applied to the extracted lane change scenario data.
Fig.~\ref{fig:data_compare} shows the distribution of extracted and generated data.

\begin{algorithm}[H]
   \caption{Lane Change Scenario Extraction}
   \label{alg:lane_change_extraction}
   \begin{algorithmic}
   \State {\bfseries Input:} Highway/OSM map, car trajectories
   \State {\bfseries Output:} Lane change scenarios
   \State \textbf{1) Lane change scenario pre-extraction:}
   \State \quad Load the highway or OSM map to get the road network structure.
   \State \quad For each trajectory:
   \State \quad \quad 1. Get the position of the ego vehicle.
   \State \quad \quad 2. Match the vehicle position with the road network to obtain lane index.
   \State \quad \quad 3. Record previous and current lane index.
   \State \quad \quad 4. If previous lane index differs from current lane index:
   \State \quad \quad \quad a) Ensure both previous and current lane indices are in the nearby lane list.
   \State \quad \quad \quad b) Record the lane change frame.
   \State \quad \quad 5. Find the vehicles on the previous and current lanes before and after the lane change.
   \State \quad \quad 6. Record the front and rear vehicles in both lanes.
   \end{algorithmic}
\end{algorithm}


   
   


\begin{figure}[th]
\begin{center}
\centerline{\includegraphics[width=1.1\columnwidth, trim=50 10 10 10, clip]{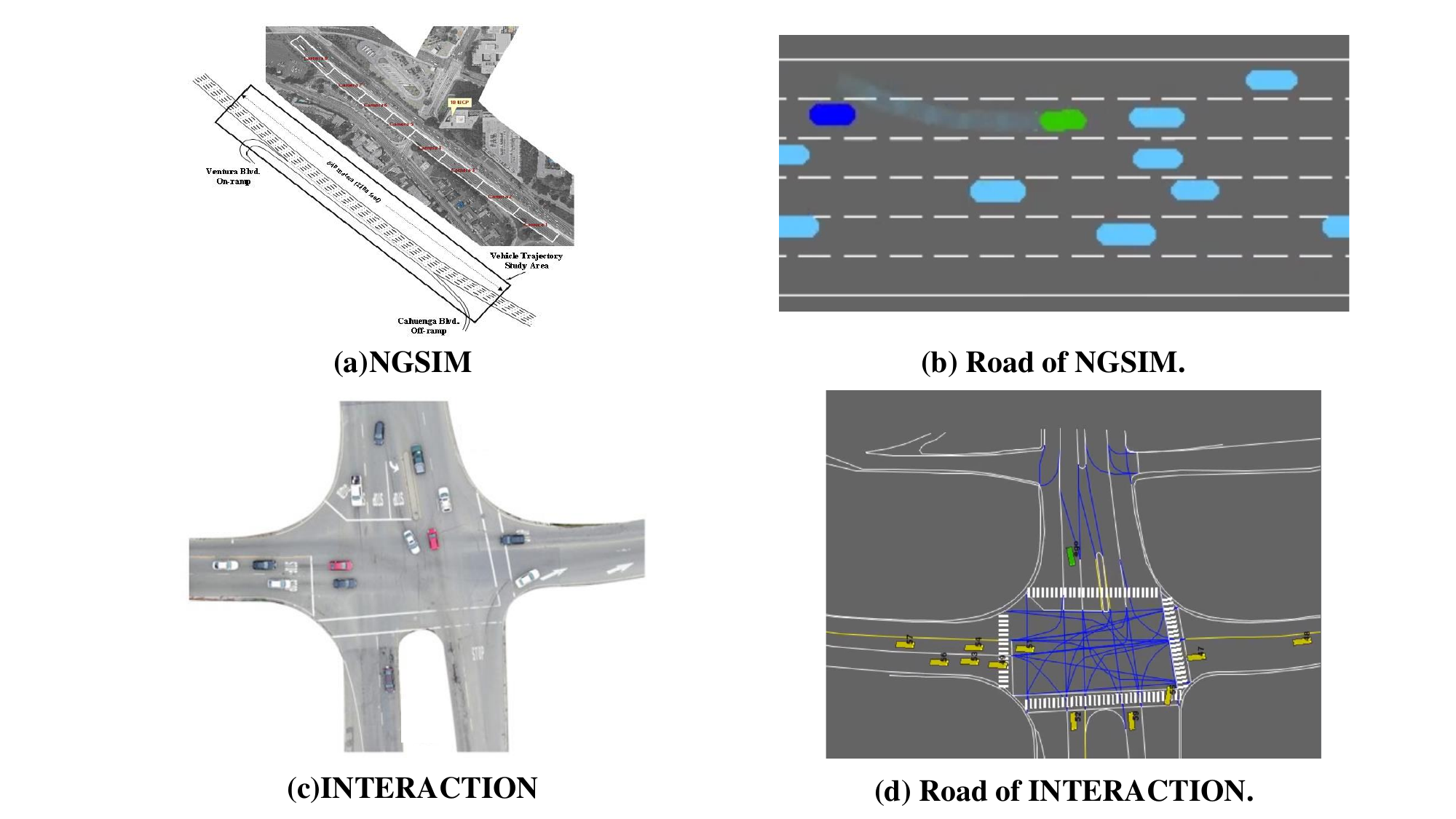}}
\caption{Constructing road structures from the real-world datasets.}
\label{fig:data_lane}
\end{center}
\vspace{-0.35in}
\end{figure}

We customized the Highway-env simulation environment constructed by Hao et al. \cite{c14} to include multi-lane highways and complex intersections, as illustrated in Fig.~\ref{fig:data_lane}.

\subsection{Model Construction and Optimization}
\subsubsection{Gail-based Generation Model}
Generative Adversarial Imitation Learning (GAIL) \cite{c28} is a method that combines Generative Adversarial Networks (GANs) with imitation learning. In driving behavior generation, GAIL effectively simulates the complex decision-making process of human drivers, thereby generating realistic driving behaviors.


    
To enhance the model's sensitivity to complex driving actions, especially on highways and intersections, the Leaky mechanism is integrated into the clipping mechanism of the PPO model~\cite{c10}, and Wasserstein Distance (W-Distance) \cite{c29, c30} is employed during training to measure naturalness of the generated behaviors. Leaky PPO allows the policy update ratio $r(\theta)$ to maintain small gradients when exceeding a predefined threshold, preventing the problem of vanishing gradient and ensuring that the model explores the policy space more thoroughly. W-Distance, a more stable metric for measuring distribution discrepancies, effectively captures the difference between the generated policy and expert behavior, especially in high-variance or sparse reward scenarios. This improves the model’s ability to learn high-risk behaviors, such as lane-changing.

Specifically, in the traditional Clipped PPO algorithm, the policy update is constrained using the ratio $r(\theta)$ to ensure the algorithm's stability. However, when the policy update ratio \( r(\theta) \) exceeds the predefined threshold (\(r(\theta) \leq 1 + \epsilon\) or \(r(\theta)  \geq   1 - \epsilon\)), gradient information is lost, leading to the policy being unable to optimize further. To optimize policy learning and avoid issues such as gradient vanishing, we implemented Leaky PPO (as shown in Fig.~4), which introduces a small positive gradient when the ratio exceeds the predefined threshold. This modification not only preserves critical gradient information to ensure continuous learning and effective adaptation to rare and challenging traffic scenarios, but also enhances policy exploration in complex spaces by relaxing the ratio-based constraint, thus alleviating pessimistic estimation issues \cite{c31}. This improvement strikes a better balance between the stability of the algorithm and the learning efficiency.

\begin{figure}[t]
\begin{center}
\centerline{\includegraphics[width=0.98\columnwidth, trim=80 110 100 60, clip]{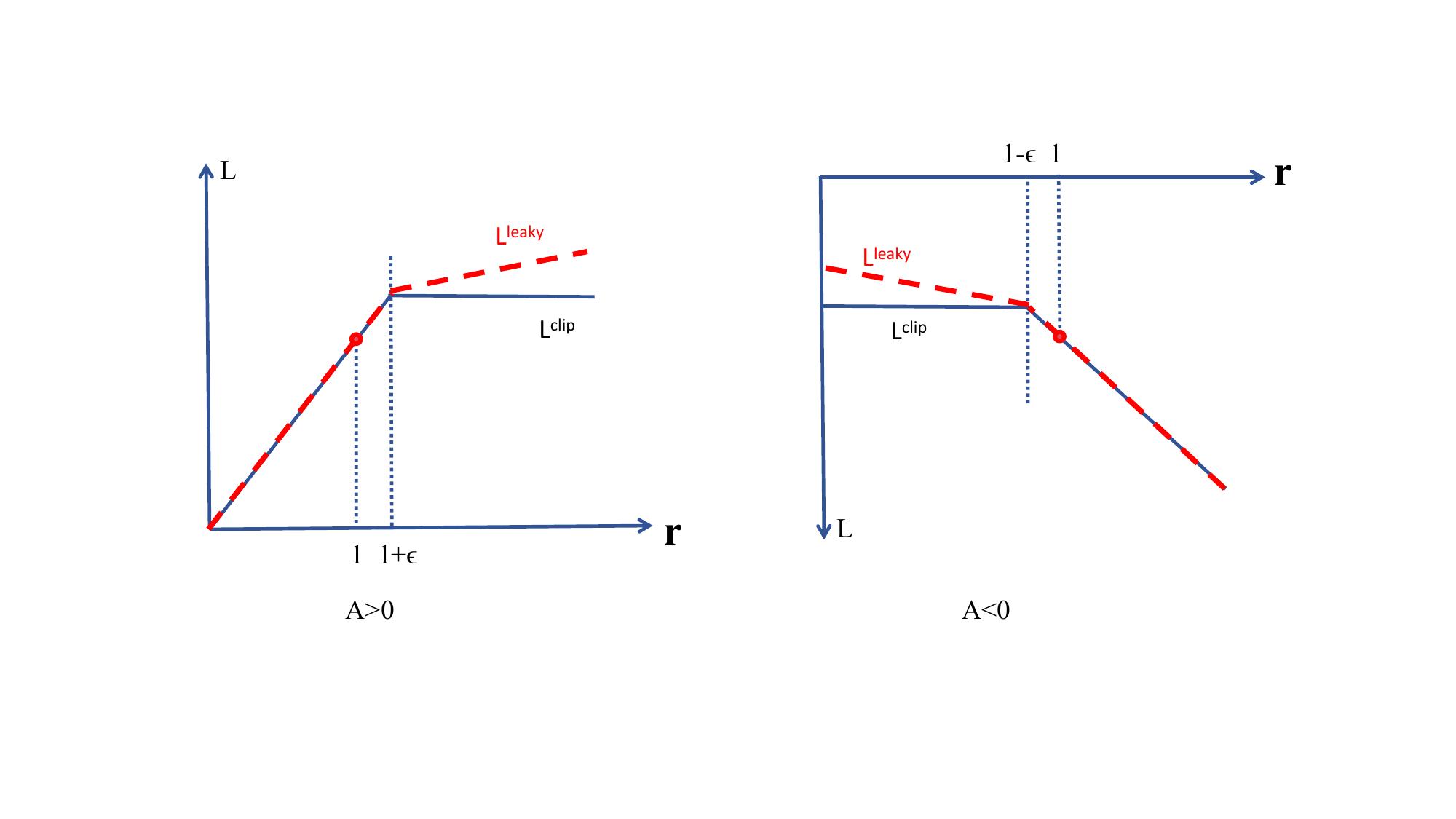}}
\caption{The plot illustrates the relationship between the objective function \( L_{\text{Leaky}}(\theta) \) and the likelihood ratio \( r \) for both positive and negative advantages, with the red point indicating the initial \( r \) value. Notably, gradients persist even within the saturation regions.}
\label{fig:leaky_ppo}
\end{center}
\vspace{-0.3in}
\end{figure}

The objective function of Leaky PPO consists of two parts: 
\textbf{1)} The standard PPO loss, which is calculated using the ratio \( r(\theta) \) and the advantage function \( \hat{A}_t \). 
\textbf{2)} When the ratio \( r(\theta) \) exceeds the predefined threshold, a small gradient is added to prevent the problem of the disappearance of the gradient.
\begin{equation}
L^{Leaky}(\theta) = E_t[min(r(\theta) \hat A_t,clip(r(\theta), l_{s,a} , u_{s,a} ) \hat A_t)]
\end{equation}

Here, \( l_{s,a} \) and \( u_{s,a} \) are the new lower and upper bounds calculated based on the threshold \( \epsilon \) and the parameter \( \alpha \), as given by the following formulas:
\begin{equation}
\text{\(l_{s,a}\)} = \alpha r(\theta) + (1-\alpha)(1-\epsilon)
\end{equation}
\begin{equation}
\text{\(u_{s,a}\)} = \alpha r(\theta) + (1-\alpha)(1+\epsilon)
\end{equation}

Where \( \alpha \) is a coefficient between 0 and 1, controlling the adjustment magnitude when the ratio exceeds the threshold.

W-Distance, which exhibits higher robustness than traditional Kullback-Leibler (KL) divergence in handling long-tail distributions and rare events, provides a stable measurement of the difference between the generated policy and the expert behavior distribution. This effectively prevents mode collapse and enhances the naturalness of the generated data.
To maximize the model's growth potential and prevent overfitting to early-stage data, i.e., to continuously explore new strategies during training, we introduce the environment reset mechanism (Resets). A common issue in DRL algorithms is the “prior bias" problem \cite{c32}, where the agent overly adapts to early environmental interactions and neglects useful evidence from later stages, resulting in poor data quality and further hindering learning performance. To address this issue, this experiment periodically reinitializes the last few layers of the neural network while retaining historical experience in the replay buffer and updating the random seed with the current training iteration. This mechanism periodically 'forgets' outdated knowledge, enabling the agent to exploit new experiences, overcome prior bias, and—by averting premature convergence—improve generalization and exploration in complex driving scenarios, thus boosting the growth of model potential.


\subsubsection{Reward Function Design}

To guide the generation of realistic, diverse, and socially adversarial behaviors, we design a composite reward function consisting of two main components: (1) a naturalness reward \( R_{\text{natural}} \) that encourages distributional similarity to expert behavior, and (2) an adversarial reward \( R_{\text{adv}} \) that promotes challenging, socially-aware interactions. The total reward is formulated as:

\begin{equation}
L = \mathbb{E} \left[ R_{\text{natural}} + \beta \cdot R_{\text{adv}} \right]
\end{equation}

where \( \beta \) balances the adversarial influence relative to the naturalness constraint.


\paragraph{Naturalness Reward.}
The naturalness reward is based on the W-Distance \( W(p_1, p_2) \) between the generated behavior distribution \( p_1 \) and expert behavior distribution \( p_2 \). This metric jointly considers the distance between distribution means and covariance structures:

\begin{equation}
\begin{aligned}
W(p_1, p_2) &= \frac{1}{B} \sum_{i=1}^B \left\| \mu_1^i - \mu_2^i \right\|_2^2 \\
\quad + &\frac{1}{B} \sum_{i=1}^B \text{Tr} \left( \Sigma_1^i + \Sigma_2^i - 2\left( \Sigma_1^i \Sigma_2^i \Sigma_1^i \right)^{1/2} \right)
\end{aligned}
\end{equation}

where \( B \) is the batch size, \( \mu \) and \( \Sigma \) represent the mean and covariance of actions in each batch. The final reward is defined as

\begin{equation}
R_{\text{natural}} = \text{clip} \left( \frac{\theta_w - W(p_1, p_2)}{\theta_w}, \ 0, \ 1 \right)
\end{equation}

The clipping operation constrains the reward within \([0, 1]\), ensuring stability and promoting behavior generation that closely aligns with expert-like trajectories.

\paragraph{Adversarial Reward.}
The adversarial reward is designed to challenge the decision-making capacity of the AV under test while remaining within plausible behavioral bounds. It consists of three components:

\begin{equation}
R_{\text{adv}} = R_{\text{SVO}} + r_{d,t} + r_{c,t}
\end{equation}

To incorporate socially-aware adversarial behavior, we extend the classical SVO reward into an adaptive formulation. The reward at time $t$ is defined as
\begin{equation}
R_{\text{SVO}}(t) = U_{\text{ego}}(t)\cos(\phi_t) + U_{\text{sv}}(t)\sin(\phi_t)
\end{equation}

where $U_{\text{ego}}(t)$ denotes the ego-centric utility and $U_{\text{sv}}(t)$ the interaction utility with surrounding vehicles. Unlike conventional SVO approaches with fixed angles, the orientation $\phi_t$ is dynamically updated during training, enabling the agent to autonomously balance egoistic efficiency and social awareness.

Each surrounding vehicle is encoded into a feature vector containing presence, position $(x,y)$, velocity $(v_x,v_y)$, orientation $(\cos\theta,\sin\theta)$, lane priority, and pairwise distance. These features, concatenated into a $V \times F$ representation for $V$ vehicles, are projected via a shared MLP and processed by a Set Transformer encoder. Multi-head self-attention (2–4 layers) captures vehicle-to-vehicle dependencies, while a pooling layer aggregates them into a global interaction embedding. The presence mask ensures robustness to a variable number of vehicles.

The social term is computed as
\begin{equation}
U_{\text{sv}}(t) = \sum_i (\beta_0 + \beta_1 p_i)\,S_i
\end{equation}

where $p_i$ is the one-hot encoding for lane priority, $\beta_0$ assigns a baseline weight to all vehicles, and $\beta_1$ serves as the importance weight for the scalar amplification of road rights. $S_i$ represents closing speed which is computed as

\begin{equation}
S_{i}=\max \left(0,-\frac{x_{i} v_{x, i}+y_{i} v_{y, i}}{r_{i}+\varepsilon}\right)
\end{equation}

 where $r_i$ denotes the relative distance between the ego and vehicle $i$, and $\varepsilon$ is a small constant added for stability to avoid division by zero. This flexible design assigns higher influence to vehicles with stronger interaction potential.

To align the reward with traffic context, the instantaneous SVO angle is updated as
\begin{equation}
\phi_t = \arctan2(U_{\text{sv}}(t), U_{\text{ego}}(t))
\end{equation}

This formulation couples the ego-progress utility with surrounding interaction effects, allowing the policy optimization process to adaptively shift between egoistic and prosocial strategies. Consequently, the SVO reward evolves from a manually tuned hyperparameter into a self-regulating mechanism, improving robustness in diverse traffic scenarios.

The overall design integrates high-dimensional per-vehicle embeddings, Set Transformer-based interaction encoding, and adaptive angle regulation within the reinforcement learning loop. This enables SCPPO to exploit both naturalistic driving priors and socially-aware adversarial objectives in a unified framework.



\begin{equation}
r_{d,t} = \text{clip} \left( 1 - \frac{ \left\| \mathbf{p}_{AV, t} - \mathbf{p}_{a, t} \right\|_2 }{ \left\| \mathbf{p}_{AV, t_0} - \mathbf{p}_{a, t_0} \right\|_2 }, \ -1, \ 1 \right)
\end{equation}

This term quantifies temporal proximity risk between the AV and adversarial agent. Positions \( \mathbf{p}_{\cdot, t} \) are sampled at current (\(t\)) and initial time (\(t_0\)) steps. Clipping ensures numerical stability during training.

\begin{equation}
r_{c,t} =
\begin{cases}
1, & \text{if collided with the AV under test} \\
0, & \text{if no collision} \\
-1, & \text{if collided with other vehicles}
\end{cases}
\end{equation}

This reward penalizes undesirable collisions with background vehicles while encouraging targeted AV interactions.

In summary, the reward function integrates Wasserstein-based distributional alignment with socially informed adversarial incentives. The use of SVO theory ensures that adversarial behaviors remain interpretable and rational, distinguishing our approach from purely random or aggressive adversarial generation methods.

\section{Experimental Process}
\subsection{Model Training}
The proposed driving-behavior generator is trained within a GAIL framework enhanced by an improved PPO optimizer. Expert-trajectory data, tailored reward shaping, and curriculum scheduling are employed, with hyperparameters tuned separately for the discriminator (GAIL) and the policy (PPO) via Bayesian search (final PPO settings in Table~\ref{ppo-parameters}).

\begin{table}[th]
\caption{PPO Training Parameters}
\label{ppo-parameters}
\begin{center}
\begin{small}
\begin{sc}
\begin{tabular}{lc}
\toprule
\textbf{parameter} & \textbf{value} \\
\midrule
Learning rate (GAIL)& 0.0003 \\
Batch size (GAIL)& 4096 \\
Number of threads (GAIL)& 24 \\
Learning rate (PPO)& 0.0002 \\
Batch size (PPO)& 2048 \\
Number of threads (PPO)& 2 \\
\(\alpha\) (Leaky PPO) & 0.01 \\
Discount factor & 0.99 \\
Max Action & \([-\pi/4, \pi/4]\) \\
w1:w2 & 6:4 \\
\(\theta_w\) (W-Distance) & 0.9 \\
Reset interval & 1000 \\
Reset network layers & 3 \\
Replay buffer capacity & 100000 \\
\bottomrule
\end{tabular}
\end{sc}
\end{small}
\end{center}
\end{table}


Training proceeds in two stages. First, filtered dataset samples are used to train the GAIL model to replicate expert driving behaviors, with the discriminator guiding the generator to simulate realistic driving behaviors. The trained GAIL model then provides supervisory signals for PPO, encouraging policies that resemble expert behavior while generating adversarial scenarios. 
In the PPO phase, the agent interacts with the customized Highway-env environment, updating policy and value networks each iteration. Leaky mechanism and W-Distance improve policy exploration and behavioral naturalness, while periodic resetting of the last three network layers mitigates prior bias and overfitting. 

\begin{figure}[th]
\begin{center}
\centerline{\includegraphics[width=1\columnwidth]{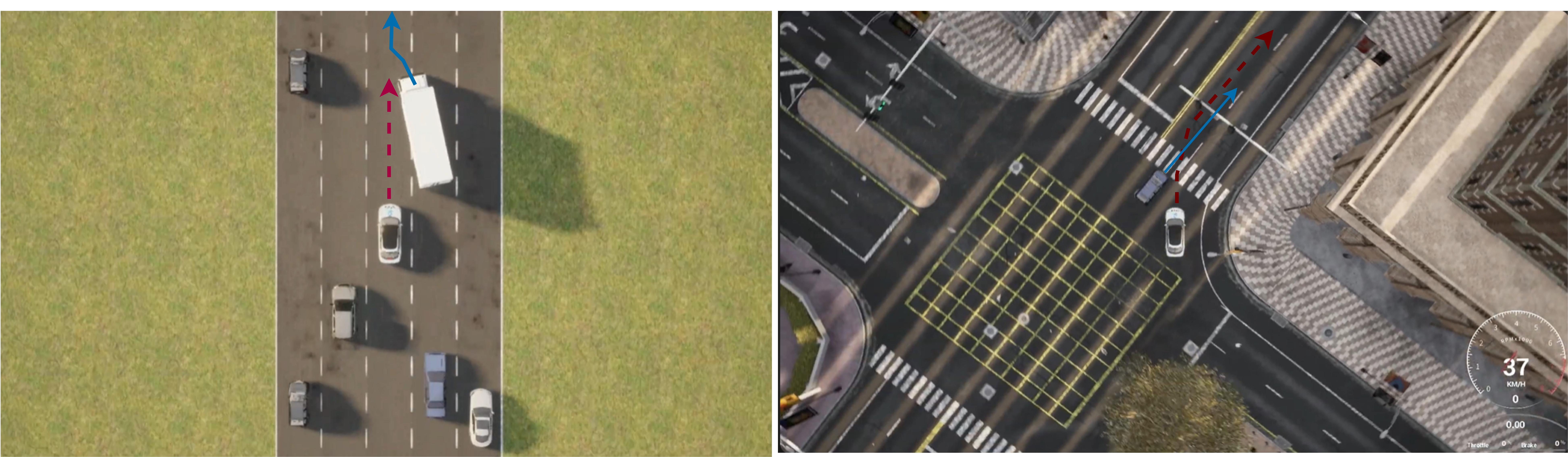}}
\caption{Visualization using CARLA for both highway and intersection.}
\label{fig:CARLA}
\end{center}
\vspace{-0.3in}
\end{figure}

This process enables the model to generate more realistic and adversarial driving scenarios, with results visualized using CARLA for both highway and intersection settings, as shown in Fig.~\ref{fig:CARLA}.

\subsection{Experimental Metrics}
1) Adversarial reward:
It evaluates the performance of driving behaviors generated in adversarial scenarios, serving as a key metric to optimizing individual rewards and rational interaction in complex traffic environments. Reflects the ability to generate dangerous yet reasonable behaviors. In addition, it combines the SVO reward with other adversarial-based reward terms. The formula is defined as follows:
\begin{equation}
\text{\(R_{adversarial}\)}= R_{SVO} + \beta R_{adv}
\end{equation}

where \( R_{\text{SVO}} \) represents the SVO reward, which reflects the agent's ability to balance intent of ego vehicle and adversarial behavior in complex interaction scenarios. \( R_{\text{adv}} \) represents other adversarial rewards, and \( \beta \) is the weight parameter for the adversarial reward.

2) Dangerousness parameter:
To quantify dangerousness under realistic and natural driving, we aggregate a collision rate with two
smoothness proxies computed over the entire episode. The final score is a convex combination
\begin{equation}
D_{\mathrm{risk}}
= w_c\, C_{\mathrm{coll}}
+ w_a\, \psi\!\left(\bar{J}_y;\,\tau_a^{L},\tau_a^{H}\right)
+ w_t\, \psi\!\left(\bar{S}_{\mathrm{traj}};\,\tau_t^{L},\tau_t^{H}\right)
\label{eq:drisk}
\end{equation}
so that $D_{\mathrm{risk}}\!\in[0,1]$ and $w_c{=}0.8,\; w_a{=}w_t{=}0.1$. The collision rate $C_{\mathrm{coll}}\!\in[0,1]$ is defined as the effective collision frequency,
i.e., the fraction of episodes in which the main vehicle and the adversarial vehicle collide:
\begin{equation}
C_{\mathrm{coll}} = \frac{N_{\mathrm{coll}}(\text{ego},\,\text{adv})}{N_{\mathrm{episodes}}}
\end{equation}

For lateral-acceleration comfort, let $y_k$ denote the lateral position at frame $k$ (sampling
period $\Delta t$). Define $v_{y,k}=(y_{k+1}-y_k)/\Delta t$, $a_{y,k}=(v_{y,k+1}-v_{y,k})/\Delta t$,
and $j_{y,k}=(a_{y,k+1}-a_{y,k})/\Delta t$. The episode-level discomfort proxy is the mean absolute
lateral jerk
\begin{equation}
\bar{J}_y = \frac{1}{T-3}\sum_{k=0}^{T-4} |j_{y,k}|\;\;(\ge 0)
\end{equation}

Trajectory smoothness is measured over the episode via the second-order finite difference of lateral position:
\begin{equation}
\bar{S}_{\mathrm{traj}} = \frac{1}{T-2}\sum_{k=1}^{T-2}\big|\,y_{k+1}-2y_k+y_{k-1}\,\big|\;\;(\ge 0)
\end{equation}

which penalizes oscillatory motions regardless of route. Both proxies are normalized to $[0,1]$ with the same mapping
\begin{equation}
\psi(x;\tau^{L},\tau^{H}) = \mathrm{clip}\!\left(\frac{x-\tau^{L}}{\tau^{H}-\tau^{L}},\,0,\,1\right)
\label{eq:norm}
\end{equation}

where $(\tau^{L},\tau^{H})$ are chosen from expert-data quantiles (e.g., 75th/95th) and can absorb the
time-scale factor $\Delta t$ if desired. This design makes the score primarily driven by the effective collision rate, while insufficient lateral comfort and poor trajectory smoothness further increase $D_{\mathrm{risk}}$ in a principled, comparable manner.

\subsection{Analysis under Different Conditions}
\begin{figure}[t]
\begin{center}
\centerline{\includegraphics[width=1.1\columnwidth, trim=10 10 10 10, clip]{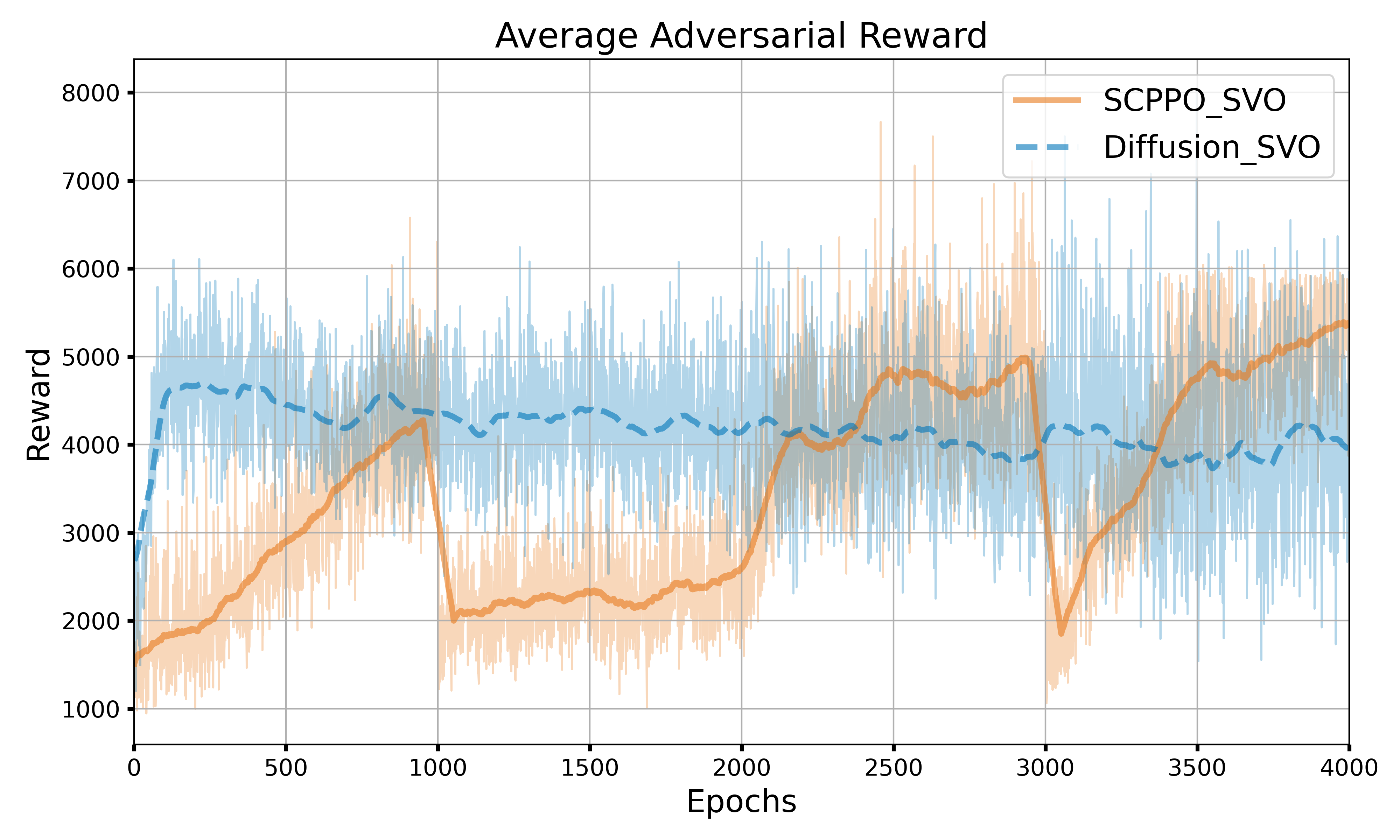}}
\caption{Comparison of the Average Adversarial Reward Between Diffusion and GAIL}
\label{fig:diffusion_gail}
\end{center}
\vspace{-0.3in}
\end{figure}

Firstly, we compare the performance of SCPPO with two baseline models: the diffusion-based and GAIL-based frameworks. The comparison is based on the average adversarial reward over 4000 training epochs, as shown in Fig.~\ref{fig:diffusion_gail}. Initially, the diffusion-based framework achieves higher adversarial rewards than SCPPO, reflecting its ability to generate more challenging scenarios at the early stages. However, this higher reward is unstable, with significant oscillations observed in the first 2,000 epochs. The performance of diffusion fluctuates as the model struggles to stabilize and improve consistently.

\begin{figure}[b]
\begin{center}
\vspace{-0.2in}
\centerline{\includegraphics[width=1.1\columnwidth, trim=10 10 10 10, clip]{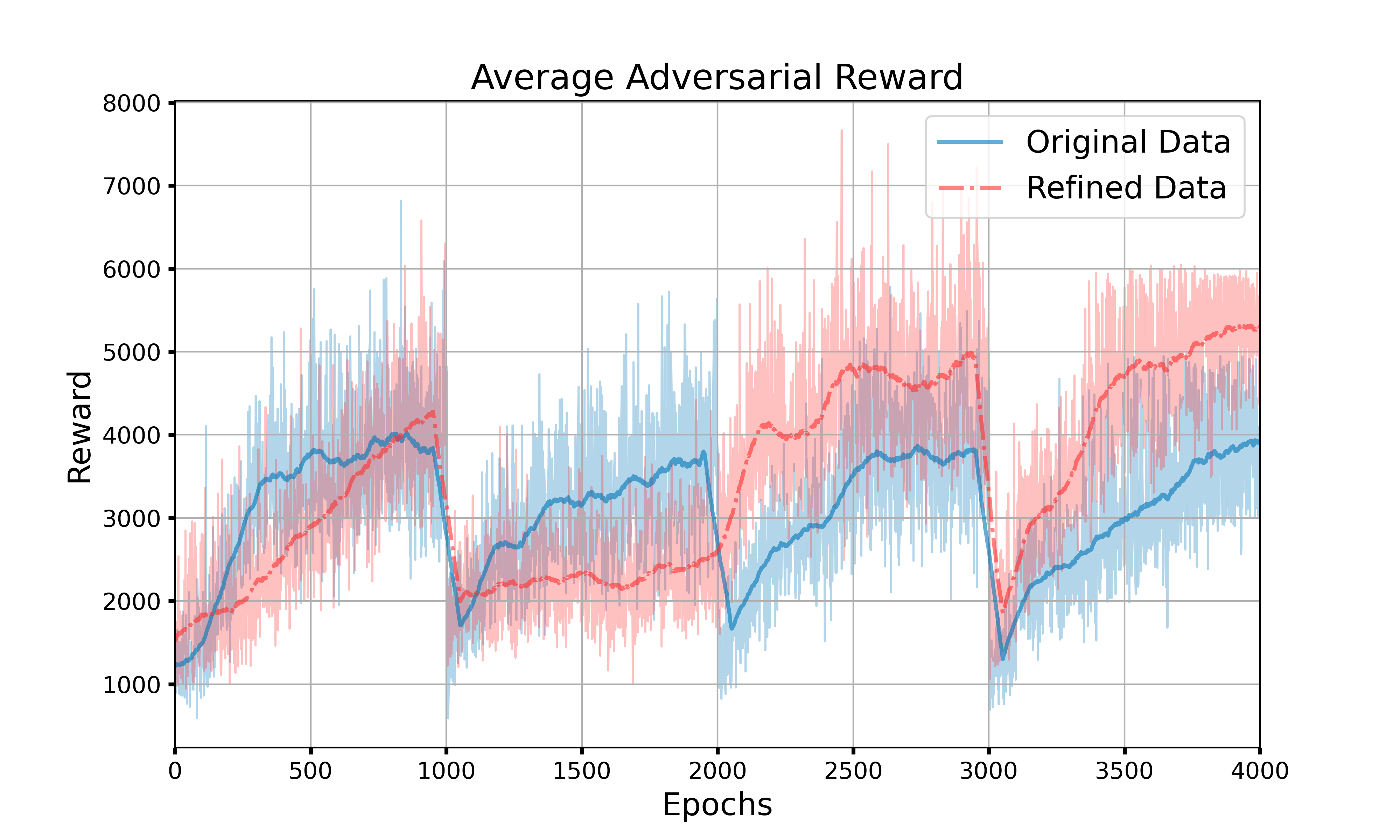}}
\caption{Comparison of the Average Adversarial Reward Between The Refined Datasets and The Original Dataset}
\label{old_avg}
\end{center}
\vspace{-0.3in}
\end{figure}

In contrast, SCPPO starts with a lower adversarial reward but shows a steady improvement over time. Despite slower initial growth between epochs 1,000 and 2,000 due to the random seed reset mechanism, SCPPO eventually surpasses Diffusion after epoch 2400, demonstrating its superior ability to generate more stable and adversarial driving scenarios.

To assess the impact of SVO on the generation of adversarial scenarios, we compare the dangerousness parameter between different SVO angles. This parameter quantifies the risk of generated behaviors, with higher values indicating more adversarial scenarios. 

\begin{table}[th]
\caption{Dangerousness Parameter for Different SVO Angles}
\label{tab:svo_dangerousness}
\begin{center}
\begin{small}
\begin{sc}
\begin{tabular}{lc}
\toprule
\textbf{SVO Angle} & \textbf{Dangerousness Parameter (\%)} \\
\midrule
SVO: -15° & 20.1 \\
SVO: -45° & \textbf{22.7} \\
SVO: 45°  & 7.8 \\
SVO: 15°  & 10.9 \\
without SVO    & 19.4 \\
\bottomrule
\end{tabular}
\end{sc}
\end{small}
\end{center}
\end{table}

As shown in Table~\ref{tab:svo_dangerousness}, 
angle values are placed in the fourth quadrant to highlight adversarial scenarios and first-quadrant angles serve as a cooperative reference for comparison. The setting SVO: -45° results in the highest dangerousness value of 22.7\%, indicating that the model generates more adversarial and risky scenarios when the SVO promotes a more competitive and antagonistic approach. On the other hand, SVO: 45° generates the least adversarial behavior, with a low dangerousness value of 7.8\%, suggesting that more cooperative SVO angles lead to less risky driving behavior. 
The baseline model, which does not integrate any SVO reward, achieves a moderate dangerousness parameter of 19.4\%, demonstrating the importance of SVO in influencing the adversarial nature of the generated scenarios. These results highlight the ability of SVO to guide the model in generating driving behaviors with varying levels of adversarial characteristics, with more extreme SVO values (e.g. -45°) leading to higher-risk scenarios.

\begin{figure}[b]
\begin{center}
\centerline{\includegraphics[width=1.1\columnwidth, trim=10 10 10 10, clip]{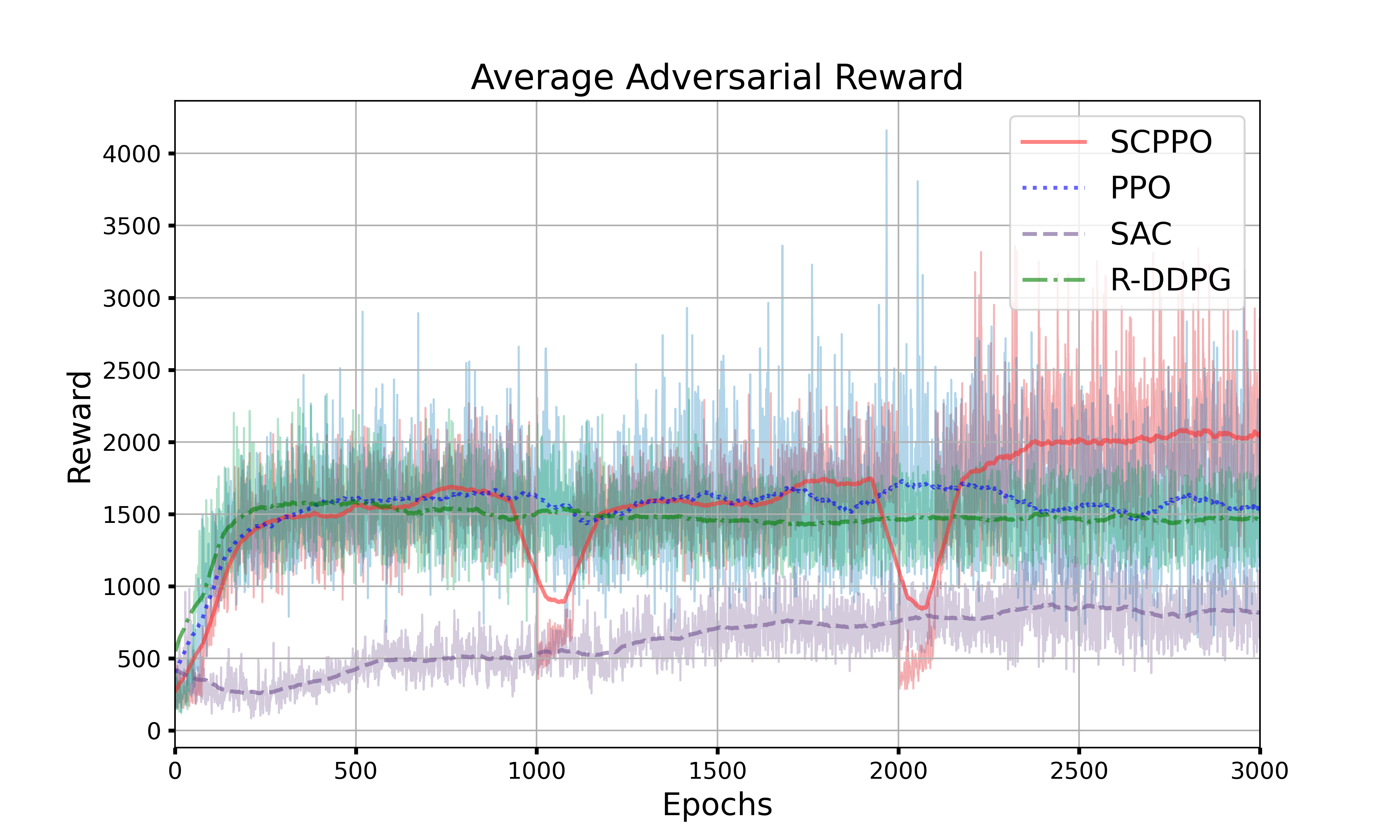}}
\caption{Comparison with Baseline Models}
\label{fig:baseline_comparison}
\end{center}
\vspace{-0.3in}
\end{figure}
Additionally, the model is trained using both the original (NGSIM and INTERACTION) and refined datasets, with comparisons based on the average adversarial reward, as shown in Fig.~\ref{old_avg}.
The experimental results show that, as training progresses, the adversarial reward of the refined dataset increases significantly more than that of the original dataset. This indicates that data selection and mining play a crucial role in enhancing the model's performance, particularly in generating adversarial driving scenarios.

\subsection{Performance Evaluation and Ablation}

The proposed SCPPO model is compared with baseline reinforcement learning algorithms (PPO and SAC) as well as the recent and relevant adversarial method R-DDPG \cite{c33}. The comparison is based on the average adversarial reward, as shown in Fig.~\ref{fig:baseline_comparison}, with all models trained on the same dataset and without SVO to ensure a fair evaluation.

\begin{figure}[t]
\vspace{-0.1in}
\begin{center}
\centerline{\includegraphics[width=1.1\columnwidth, trim=10 10 10 10, clip]{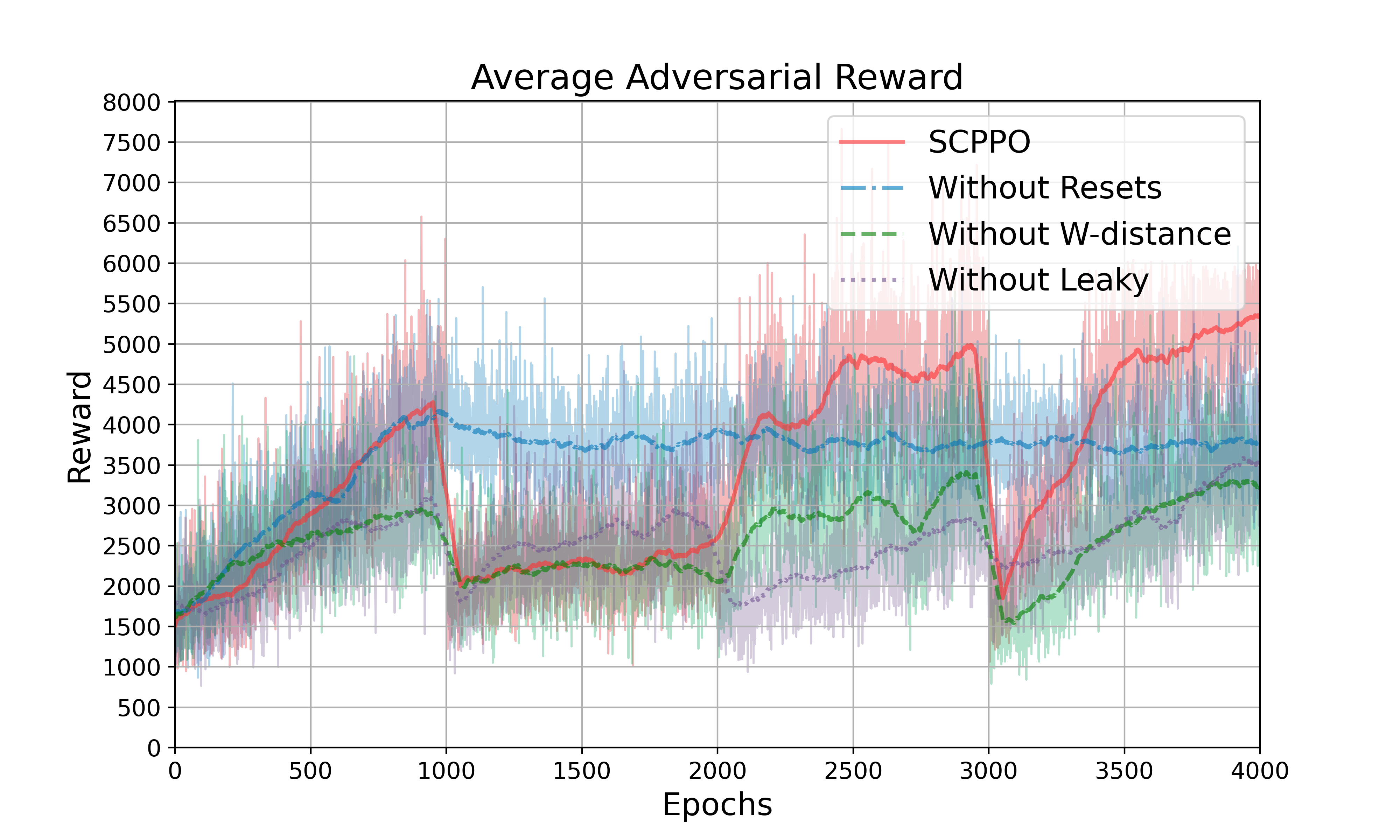}}
\caption{Ablation Study (Comparison of Different Mechanisms: Leaky Mechanism, Resets, W-Distance)}
\label{SCPPO}
\end{center}
\vspace{-0.3in}
\end{figure}

Experimental results demonstrate that SCPPO consistently outperforms PPO, SAC, and R-DDPG, in terms of average adversarial reward after 2,200 epochs. SCPPO shows a rapid increase in reward starting from the early stages of training, stabilizing at a higher level compared to the other models. Although SAC exhibits slower convergence and attains lower rewards throughout training, SCPPO demonstrates more effective learning, achieving higher rewards, and maintaining them at a stable level much earlier in the process.
In contrast, PPO and R-DDPG, although they show relatively stable performance, still lag behind SCPPO in terms of reward progression. The combination of sensitivity and continuous learning of SCPPO allows it to generate more challenging and adversarial driving behaviors. This advantage is especially pronounced in the generation of high-risk driving behaviors, where the SCPPO model achieves higher reward levels, demonstrating a stronger ability to generate desired driving scenarios.

 To verify the contribution of each improvement module to the performance of the final model, ablation experiments are designed to analyze the role of these modules and their influence on each other by removing the Leaky mechanism, Resets and the W-Distance one by one.

The ablation results in Fig.~\ref{SCPPO} demonstrate that eliminating any of these modules consistently lowers the adversarial reward, highlighting the critical role of the Leaky mechanism, Resets, and W-Distance in enhancing SCPPO’s performance.

\section{Conclusion}
This study developed a refined rule-based data mining process based on existing open-source datasets. Dangerous interaction behaviors are identified in the NGSIM and INTERACTION datasets, and a driving behavior generation model based on the improved GAIL framework is proposed for such data, focusing on the generation of simulation data for lane change behaviors in complex traffic scenarios. By incorporating the Leaky mechanism, W-Distance and Resets into the PPO algorithm, and integrating SVO into the reward function, the model demonstrates significant advantages in capturing and generating rare and complex driving behaviors.

Experimental results show that the proposed model outperforms the baseline models in key metrics such as dangerousness parameter, and adversarial reward, exhibiting higher sensitivity and adaptability. It is capable of generating more natural and reasonably adversarial driving behaviors based on the refined data we extracted. The generated data is further refined through a data processing pipeline to ensure greater validity. Since the reinforcement learning algorithm in GAIL learns policies rather than data distributions, the model can be readily tested for robustness in unseen scenarios, particularly in handling long-tail lane-change cases. For augmenting the distribution of small sample behaviors in the dataset, our framework achieves a conversion success rate of approximately 13\%, providing a foundation for subsequent training on long-tail lane change scenarios.
Despite the significant results of this study, several directions remain for further exploration. For example, more refined mining of open-source datasets with additional modalities, such as Waymo and nuScenes, needs further investigation.

\end{document}